\documentclass{article}

\usepackage{amssymb}  
\usepackage{amsmath}
\usepackage{arxiv}
\usepackage[utf8]{inputenc} 
\usepackage[T1]{fontenc}    
\usepackage{hyperref}       
\usepackage[normalsize]{caption} 
\usepackage{floatrow} 
\usepackage{float}
\usepackage[position=bottom, font=Large]{subfig} 
\usepackage{bigdelim}
\usepackage{booktabs} 
\usepackage{hhline} 
\usepackage{longtable}
\usepackage{supertabular}
\usepackage{multirow}
\usepackage{tabu} 
\usepackage{tabularx} 
\usepackage{graphicx}
\usepackage{comment}
\usepackage{multirow}
\usepackage{graphicx}
\usepackage[normalem]{ulem}
\useunder{\uline}{\ul}{}
\usepackage{mathtools} 

\title{\textit{META4}: Semantically-Aligned Generation of Metaphoric Gestures using Self-Supervised Text and Speech Representations}

\date{} 					

\author{
{
\hspace{1mm}Mireille Fares}\\
    ISIR, STMS \\
	Sorbonne University\\
	Paris, France\\
	\And
 {
 \hspace{1mm}Catherine Pelachaud} \\
    ISIR, CNRS \\
	Sorbonne University\\
	Paris, France\\
 \And
  {
  \hspace{1mm}Nicolas Obin}\\
    STMS \\
	Sorbonne University\\
	Paris, France\\}

\renewcommand{\headeright}{ }
\renewcommand{\undertitle}{ }
\renewcommand{\shorttitle}{META4}

\hypersetup{
pdftitle={A template for the arxiv style},
pdfsubject={q-bio.NC, q-bio.QM},
pdfauthor={David S.~Hippocampus, Elias D.~Striatum},
pdfkeywords={First keyword, Second keyword, More},
}

\begin{document}

\maketitle

\begin{abstract}
Image Schemas are repetitive cognitive patterns that influence the way we conceptualize and reason about various concepts present in speech. These patterns are deeply embedded within our cognitive processes and are reflected in our bodily expressions including gestures. Particularly, metaphoric gestures possess essential characteristics and semantic meanings that align with Image Schemas, to visually represent abstract concepts. The shape and form of gestures can convey abstract concepts, such as extending one's forearm and hand or tracing a line with hand movements to visually represent the image schema of "PATH". Previous behavior generation models have primarily focused on utilizing speech (acoustic features and text) to drive the generation model of virtual agents. They have not considered key semantic information as those carried by Image Schemas to effectively generate metaphoric gestures. To address this limitation, we introduce \textit{META4}, a deep-learning approach that generates metaphoric gestures from both speech and Image Schemas. Our approach has two primary goals: (1) computing Image Schemas from input text to capture the underlying semantic and metaphorical meaning, and (2) generating metaphoric gestures driven by speech and the computed image schemas. We make two key contributions: (1) BERTIS, a model that computes Image Schema tags from text input, and (2)  META4, which makes use of BERTIS to model and synthesize the corresponding metaphoric gestures from speech and the generated Image Schemas. To the best of our knowledge, our approach is the first method for generating speech-driven metaphoric gestures while leveraging the potential of Image Schemas. We present evaluation studies to demonstrate the effectiveness of our approach and highlight the importance of both speech and image schemas in modeling metaphoric gestures. Link to code, data and videos: \textit{https://github.com/mireillefares/META4/}.
\end{abstract}

\section{Introduction}
In the realm of human communication, our cognitive processes play a vital role in shaping the way we conceptualize and reason about various ideas and concepts. One significant aspect of this cognitive influence is manifested through repetitive patterns known as "\textit{Image Schemas}". These patterns, as described by Johnson et al. \cite{johnson2005philosophical}, exert a profound impact on our understanding and expression of concepts within speech. They are deeply ingrained in our cognitive processes and find expression in our bodily movements and gestures, as observed by Cienki \cite{cienki2008metaphor}. Of particular interest are \textit{metaphoric gestures}, which are gestures that symbolically represent a concept, object, or event \cite{mcneill1982conceptual}. These gestures possess inherent characteristics and semantic meanings that align with \textit{Image Schemas}. They serve as visual representations of abstract concepts \cite{kendon2004gesture}, employing specific shapes and forms to convey complex ideas. The conveyed image is a visual representation that is associated with something concrete and actionable in the world. \textcolor{black}{For example, one can sweep his/her flat hand through space, or trace a surface, to visually represent the image schema "SURFACE". When a speaker discusses the promotion of an individual in an organization, he/she may use a metaphoric gesture such as raising his/her hand upward to symbolize the promotion}. Despite the significant role that Image Schemas and metaphoric gestures play in conveying abstract concepts, previous gesture generation models \cite{ravenet2018automating, kucherenko2020gesticulator, yoon2019robots, yoon2020speech, fares2020towards, fares2022zero} for virtual agents have primarily focused on using speech acoustic features alone, or in combination with text semantics, to drive the gesture generation process. However, these generative models do not compute and capture the gesture shapes, nor make use of the rich semantic information conveyed by Image Schemas \cite{grady2005image} to synthesize metaphoric gestures.

To address this limitation, \textcolor{black}{we leverage recently developed self-supervised representations in text (BERT \cite{devlin-etal-2019-bert}) and audio (AST \cite{gong2021ast}) to enhance the input representations for multimodal neural networks for gesture synthesis, including metaphorical gestures. More specifically, } we propose \textit{META4} (short for \textit{METAPHOR}), a deep-learning approach that generates metaphoric gestures using both speech and Image Schemas. Our approach encompasses two primary objectives: (1) computing Image Schemas from input text to capture the underlying semantic and metaphoric meaning, and (2) generating metaphoric gestures driven by both speech and the computed Image Schemas. Motivated by these objectives, we make two key contributions. First, we present \textit{BERTIS}, a model designed to compute Image Schema tags from textual input. Second, we propose \textit{META4}, which builds upon \textit{BERTIS} to model and synthesize metaphoric gestures based on speech input and the generated \textit{Image Schemas}. 

To the best of our knowledge, our approach represents the first method for generating speech-driven metaphoric gestures while leveraging the potential of \textit{Image Schemas}. We conduct evaluation studies to demonstrate the effectiveness of our approach and show the importance of considering both \textit{speech} and \textit{Image Schemas} in modeling metaphoric gestures. Our contributions can be listed as follows:
\begin{enumerate}
    \item \textcolor{black}{We utilize the advancements in self-supervised representations in text (BERT \cite{devlin-etal-2019-bert}) and audio (AST \cite{gong2021ast}) to enhance the input representations of multimodal neural networks for synthesizing gestures including metaphorical gestures.}
    \item We propose \textit{\textbf{META4}}, a novel speech-driven and semantically-aligned multimodal Transformer-based approach that combines speech and Image Schemas modalities to generate metaphoric gestures.
    \item We propose \textbf{\textit{BERTIS}}, an Image Schema computational model that classifies an input text into an Image Schema class.
\end{enumerate}
\textcolor{black}{This paper is organised as follows. We start by discussing some background and a review of the existing gesture generation approaches works. We then explain and introduce the architecture of \textbf{\textit{META4}} and \textbf{\textit{BERTIS}}. Finally, we present objective evaluations and discuss the results.}

\section{Background and Related works}
\subsection{Image Schemas}
\textcolor{black}{The concept of Image Schema was first introduced in 1987 in Johnson's book ``The body in the mind''\cite{johnson2013body}. The original definition was set as follows: ``An image schema is a recurring, dynamic pattern of our perceptual interactions and motor programs that gives coherence and structure to our experience'' \cite{johnson2013body}. The idea is to define a parallel between how the body perceives and acts, and how the mind understands and knows; implying that sensory-motor capacities are recruited for abstract thinking \cite{johnson2005philosophical}.
Such patterns regroup all types of cues accumulated since childhood that are then flexibly categorized by language, allowing to put words on general and metaphoric concepts \cite{langacker1987foundations}.
To illustrate Image Schemas, let us take an example: CONTAINMENT that 
can be found in multiple expressions, such as ``Putting an idea inside someone's head''. CONTAINMENT emerges structurally from the bodily experience of either seeing something going into something else, putting something into something else, or going oneself into something. Studies in cognitive linguistics suggest that over two dozen different image schemas appear regularly in people’s everyday thinking, reasoning, and imagination }\cite{johnson2013body,lakoff1987image,gibbs2006image}. For example, Cienki \cite{cienki2005image} suggests that the image schemas CONTAINER, CYCLE, FORCE, OBJECT, PATH are reliably used to categorize gestures observed from natural conversations \cite{cienki2013image}. 


\textcolor{black}{
Several studies have highlighted the links between image schemas and gestures. Gestures and speech arise from a common cognitive process \cite{kendon2004gesture}. Gestures embody thoughts. They do not always convey the same information; they may complement each other, or even one may substitute for the other one. Gesture features, such as hand shape or wrist motion, are commonly found to illustrate image schemas. For example, the image schema PATH is often marked by the linear trajectory of the hand \cite{williams2008gesture,cienki2013image}, while the CYCLE Image Schema can be aligned with a repeated circular movement of the hand \cite{ladewig2011putting}. Image schemas can also be combined. For example, a container can be lifted, filled in, or put aside; it can be augmented with adjectives (small, hard, large, etc) that will be reflected in the gesture shape and trajectory  \cite{antonova2020container}
}

\subsection{Gesture Generation Approaches}
Previously, there have been various approaches proposed for gesture generation. The earliest approaches
\cite{pelachaud1996generating, cassell, pelachaud2002embodied, kopp2006towards} were rule-based, relying on predefined correspondences between patterns of human communication and behavior. However, these rule-based approaches have limitations in terms of requiring significant human effort to determine the rules, resulting in limited and repetitive gestures.

To overcome these limitations, researchers turned to statistical approaches \cite{kipp2005gesture, kipp2001anvil, neff2008gesture, bergmann2009gnetic, marsella2013virtual}, which synthesize gestures based on statistics from a corpus of human non-verbal behavior. While statistical models mitigated some of the limitations of rule-based approaches, they still suffered from a lack of diversity and variability in the generated gestures.

More recently, learning-based models, often referred to as data-driven models, have been proposed. These models are trained on large amounts of data and utilize machine learning algorithms. Gesture generative models based on sequential generative parametric models like Hidden Markov Models (HMMs) \cite{hofer2007automatic}, Recurrent Neural Networks (RNNs) \cite{wang2021audio2head, haag2016bidirectional}, and Dynamic Bayesian Networks (DBNs) \cite{mariooryad2012generating, sadoughi2019speech} have been used to generate head motion from speech. Generative Adversarial Networks (GANs) \cite{karras2017audio, vougioukas2019realistic} have also been employed to produce facial gestures from speech.

However, a common limitation of many of these works is that they primarily rely on a single modality of human communication, typically speech, as input. Some approaches use text transcriptions of language to synthesize gestures, but gestures are influenced by both speech prosody and language, including facial, hand, and body gestures. For instance, Ishi et al. \cite{ishi2018speech} proposed a text-based approach for generating gestures to control a humanoid robot. Their method translated text into gesture motions by associating words with concepts, concepts with gesture categories (iconic, metaphoric, deictic, beat, emblem, and adapter), and finally, gesture categories with specific gesture motions. Other works \cite{chiu2015predicting, kucherenko2020gesticulator, yoon2020speech, ahuja2020no, fares2020towards-long, fares2022transformer, fares2023zero} have attempted to fuse both modalities, incorporating speech prosody and language, for gesture synthesis. While such approaches capture well the rhythm and fluidity of gesture motion they lack of semantic expressivity.

An earlier work by Ravenet et al. \cite{ravenet2018automating} developed an approach for generating metaphoric gestures by extracting metaphorical properties from input speech using Behavior Markup Language (BML) \cite{kopp2006towards}. Their method involved extracting image schemas using WordNet. Then speech audio and gestures are synchronized through BML annotations; BML configures as well various aspects of gestures, such as hand shape, movement, and wrist orientation. Such a method allows conveying effectively  the intended representational meaning during behavior realization. However this model was not trained on real data.

Thus, a common limitation of the latest works is that they do not compute and capture the shape of gestures nor utilize the rich semantic information conveyed by Image Schemas \cite{grady2005image} to synthesize metaphoric gestures. Our aim is to cover this gap.

\section{META4: Metaphoric Gesture Generation}
\begin{figure}[h]

\centering
\includegraphics[width=6cm, height=12cm]{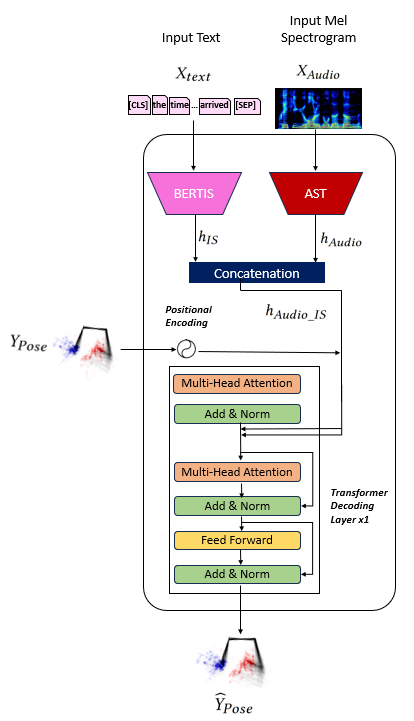}
\caption{Overview of \textbf{\textit{META4}}, a speech-driven and semantically-aligned multimodal approach for metaphoric gesture synthesis. \textbf{\textit{META4}} exploits both speech and image schema modalities as inputs, to capture the semantic meaning in speech and Image Schemas.}
\label{META4}
\end{figure}
\textcolor{black}{In this work, and based on existing neural architectures for generating gestures from speech, we focus on more complex tasks such as generating metaphorical gestures. }
We introduce \textit{\textbf{META4}}, a speech-driven and semantically-aligned multimodal Transformer-based approach that combines speech and Image Schemas modalities to generate communicative gestures paying attention to their shape and motion. \textit{\textbf{META4}} is designed as a multimodal system that leverages both speech and Image Schemas modalities to capture the underlying semantic and metaphorical meaning of the input speech and text more effectively. The Image Schemas modality is computed from the input text by employing \textit{\textbf{BERTIS}}, a pre-trained Image Schema computation model that we have independently developed and trained. Figure \ref{META4} illustrates the overall architecture of \textit{\textbf{META4}}. 
\subsection{Problem Positioning}
\begin{figure*}[]
\centering
\includegraphics[width=\textwidth]{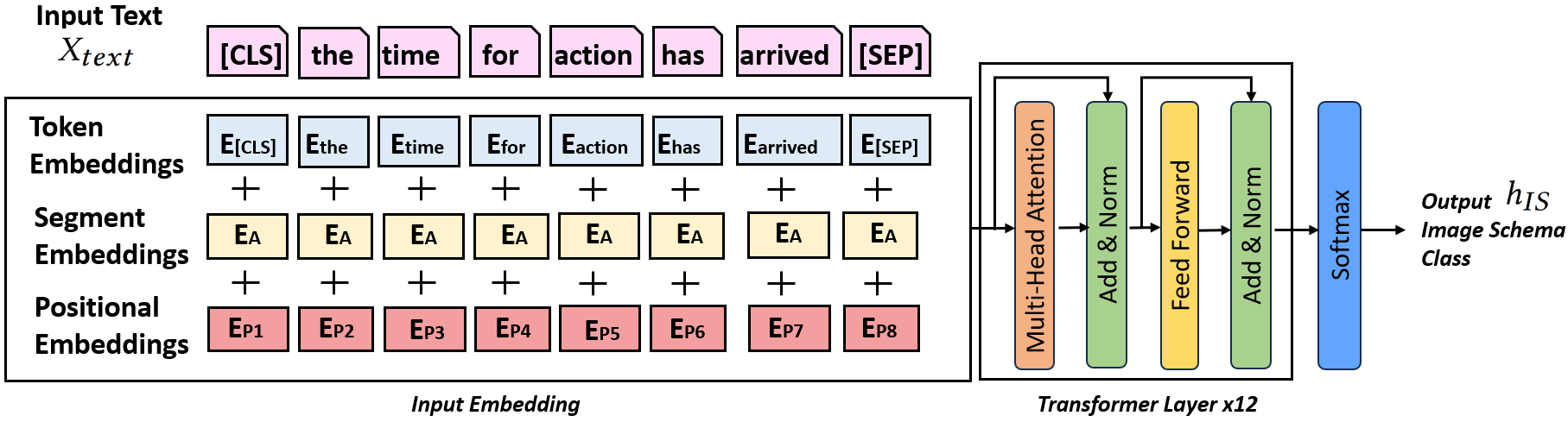}
\caption{Overview of \textbf{BERTIS} architecture, a model for Image Schema Computation. We fine-tune the \textit{BERT Base Cased model} to classify the input text $X_{text}$ into an Image Schema class $h_{IS}$.}
\label{bertis}
\end{figure*}
The goal is to generate the upper body gestures, represented by 2D poses, by utilizing both speech and Image Schemas. More specifically, we propose a novel Transformer-based architecture that effectively captures the interplay between speech, Image Schemas, and gestures. Our approach is based on the following hypothesis:
\begin{enumerate}
    \item Image Schemas are powerful conveyors of semantic information and metaphorical meaning, which share common meanings with metaphoric gestures. Complex and abstract concepts can be conveyed by employing specific gestural shapes and forms which are directly linked to the characteristics and semantic meanings conveyed by Image Schemas. 
    \item By jointly modeling Image Schemas and speech acoustic features, we can enhance the generation of metaphoric gestures by capturing more precisely gestures' shape and motion, allowing a better representation of the relationships between acoustic, metaphorical, and semantic features.
\end{enumerate}
To implement these assumptions, we propose an architecture for encoding a speaker's speech represented by his/her audio and text information and synthesizing the corresponding upper body gestures, including metaphoric gestures. Our approach includes two primary objectives:
\begin{enumerate}
    \item Computing Image Schemas from input text to capture the underlying semantic and metaphorical meaning.
    \item Generating metaphoric gestures driven by both speech and the computed Image Schemas.
\end{enumerate}
The network processes input speech and text data at the segment level \textbf{\textit{S}}, where each segment \textbf{\textit{S}} consists of 64 frames representing $4.266$ seconds that also includes silence. For each \textbf{\textit{S}} the network takes as input the speaker's Mel spectrogram ($X_{audio}$) and the corresponding text ($X_{text}$, sequence of words corresponding to the segment \textbf{\textit{S}}). For each \textbf{\textit{S}}, the output of the network is the generation of the corresponding upper-body gestures represented by 2D poses ($\widehat{Y}_{Pose}$).

\subsection{Neural Formulation}
In the subsequent subsections, we provide a detailed explanation of each module in \textit{\textbf{META4}}.
\subsubsection{Background - BERT}
BERT (Bidirectional Encoder Representations from Transformers) \cite{kenton2019bert} is a language representation model that has significantly advanced the field of natural language processing. It is built on the Transformer encoder \cite{vaswani2017attention} architecture, allowing it to capture contextual information from both the preceding and following words or tokens. In addition to standard tokens, BERT incorporates special tokens like [CLS] (classification) and [SEP] (separator) for specific purposes. The [CLS] token represents the overall representation of the input sequence and is commonly used for classification tasks. It enables BERT to generate a fixed-size representation that encapsulates the entire meaning of the input. The [SEP] token is used to separate different sentences within a single input sequence, facilitating sentence boundary understanding and inter-sentence relationship modeling. To construct the input representation, BERT sums three types of embeddings: token embeddings, segment embeddings, and position embeddings. Token embeddings capture the semantic meaning of individual words or tokens, segment embeddings differentiate between different segments or sentences, and position embeddings provide positional information within the input sequence. BERT comes in two versions: \textit{BERT base} and \textit{BERT large}. In this study, we utilize \textit{BERT base}, which consists of $12$ Transformer blocks, $12$ attention heads, and a hidden layer size of $768$.

\subsubsection{Background - BERT Fine-Tuning}
During the fine-tuning process of BERT, the parameters of the Transformer blocks, attention heads, and hidden layers, along with additional task-specific layers, are fine-tuned end-to-end. This allows the model to adapt its learned representations to the specific downstream task, leveraging both the pre-training on large-scale language data and the task-specific data during fine-tuning.

\subsubsection{Image Schema Computation Model ($BERTIS$)} 
To compute the Image Schema class $h^i_{IS}$ for a given input text $X_{text}$, we employ \textit{BERTIS}, a pre-trained model that we developed and trained independently from \textit{META4}. Specifically, we fine-tune the \textit{BERT Base Cased model} \cite{devlin2018bert} for the task of classifying the input text $X_{text}$ into an Image Schema class $h^i_{IS}$, where $i$ represents the class label. The set of image schema labels considered in this study consists of $14$ classes, as originally proposed by \cite{wachowiak2022systematic}. We refer the readers to the appendix for a comprehensive list of these Image Schemas. The fine-tuning is done by adding a fully connected dense layer on top of the output layer of \textit{BERT} and re-training the entire model using 80\% of the Image Schema Corpus introduced by \cite{wachowiak2022systematic}. 10\% of this corpus was used for validation, and 10\% for testing. As illustrated in Figure \ref{bertis}, \textit{BERTIS} takes the input text $X_{text}$ corresponding to the segment level \textbf{\textit{S}}. It computes an input embedding by adding up the token, segment, and position embeddings of each token. The computed input embedding is then given as input to $12$ Transformer layers. A SoftMax activation is added on top of the model to predict the likelihood of the output class $h^i_{IS}$, which can therefore be written as follows:
\begin{equation} \label{eqn1}
	h^i_{IS} =  E_{BERTIS}(X_{text})
\end{equation}

\subsubsection{Audio Encoder ($E_{audio}$)}
The speech modality is encoded using $E_{audio}$, the pre-trained Audio Spectrogram Transformer (AST) \textit{base384} model proposed by Gong et al. \cite{gong2021ast}. AST operates by first splitting the input Mel spectrogram $X_{audio}$, which has $128$ frequency bins, into a sequence of $16x16$ patches with overlap. These patches are then linearly projected into a sequence of 1D patch vectors, which are augmented with positional embeddings. A special [CLS] token is appended to this sequence. The resulting sequence is then input to a Transformer Encoder. Originally designed for audio classification, we modify the AST for our purposes by removing the linear layer with a sigmoid activation function at the output of the Transformer Encoder, as we do not require a classification task. Instead, we use the output of the Transformer Encoder's [CLS] token as the representation of the Mel spectrogram. The Transformer Encoder in our modified AST has an embedding dimension $d_{model}$ equals to $64$, $12$ encoding layers, and $12$ attention heads. The output vector $h_{audio}$ can therefore be written as follows:
\begin{equation} \label{eqn2}
	h_{audio} =  E_{audio}(X_{audio})
\end{equation}

\subsubsection{2D Upper Body Gesture Generator ($G_{Pose}$)}
The generated audio and image schema encoding vectors $h_{audio}$ and $h^i_{IS}$ are then concatenated together. The resulting vector $h_{Audio\_IS}$ can therefore be written as follows:
\begin{equation} \label{eqn3}
	h_{Audio\_IS} =  \left[E_{audio}(X_{audio}), E_{BERTIS}(X_{text})\right]
\end{equation}
The vector $h_{Audio\_IS}$ is subsequently provided as input to a Transformer decoder with a single decoding layer. To incorporate positional information, a positional encoding function is applied to the ground truth sequence of upper-body gestures $\widehat{Y}_{Pose}$, which is then fed into the Transformer decoder during training time. The Transformer decoder generates a probability distribution over the sequence of upper-body gestures $\widehat{Y}_{Pose}$ that corresponds to the segment \textbf{\textit{S}}. The resulting output vector $\widehat{Y}_{Pose}$ can therefore be written as follows:
\begin{equation} \label{eqn4}
	\widehat{Y}_{Pose} =  G_{pose}(h_{Audio\_IS})
\end{equation}

\section{Experimental Evaluations}
We describe the experimental evaluations in this section, and more specifically the datasets used for training, testing, and validating both \textbf{\textit{BERTIS}} and \textbf{\textit{META4}}; as well as the objective evaluation we conduct to assess our approach objectively, and post-hoc visualization of the generated 2D upper body poses.
\subsection{Material and Model setups. }
\subsubsection{PATS Corpus}We train \textbf{\textit{META4}} on the \emph{PATS Corpus} \cite{ahuja2020style, ginosar2019learning} which comprises various modalities, including 2D upper-body joint keypoints, aligned with Mel spectrogram, and Bert embeddings, from $25$ speakers. The speakers in the corpus are categorized into different groups, namely $15$ talk show hosts, 5 lecturers, $3$ YouTubers, and $2$ televangelists. Each speaker exhibits a unique communication style, contributing to lexical and gesture diversity within the corpus. The corpus consists of a total of $251$ hours of data, with $84,000$ intervals, and an average duration of $10.7$ seconds per interval. The standard deviation for interval duration is $13.5$ seconds. Each interval corresponds to an utterance containing 64 timesteps and corresponding to \textbf{\textit{S}}=$4.666$ seconds. Despite the inclusion of finger data in the PATS Corpus, we have made the decision not to incorporate finger modeling in our work. This choice is based on the observation that the quality of the extracted finger data is highly noisy and lacks accuracy. Instead, our focus lies in modeling and predicting the 2D joints of the upper body and arms, utilizing 11 joints specifically to represent these regions.
\subsubsection{Image Schemas Corpus. }We fine-tune the \textbf{\textit{BERTIS}} model using a corpus proposed by Wachowiak et al. \cite{wachowiak2022systematic}. This corpus is specifically designed to aid researchers in classifying natural language expressions into image schemas, and it contains examples from the image schema literature. The annotation data encompass samples in various languages, but in this study, we focus solely on the English samples. The English subset consists of 1994 utterance samples along with their corresponding image schema labels. During the training of \textbf{\textit{BERTIS}}, we perform data oversampling to address the class imbalance and compensate for the limited number of samples used for fine-tuning \textbf{\textit{BERT}}. This oversampling technique helps improve the classification accuracy of the model.

\subsubsection{\textbf{\textit{META4}} Testing. }We evaluate the performance of \textbf{\textit{META4}} following a testing protocol that includes two conditions: the "Seen Speaker" condition and the "Unseen Speaker" condition. In the "Seen Speaker" condition, we assess how accurately our model can generate gestures when presented with speakers that were included in the training data. The "Unseen Speaker" condition evaluates our model's ability to generalize its predictions to speakers that were not encountered during the training phase. PATS database already provides predefined train, validation, and test sets for each speaker. In our experiments, we train our model using data from $16$ PATS speakers. For the "Seen Speaker" condition, we use the test sets specifically designated for the $16$ PATS speakers as our evaluation dataset. For the "Unseen Speaker" condition, we select $6$ speakers and utilize their respective test sets to conduct our experiments.

\subsection{Objective Evaluation}
We conduct an objective evaluation of the performance of our \textbf{\textit{META4}} model by considering two crucial aspects: (1) the \textit{distance} between the predicted gestures $\widehat{Y}_{Pose}$ and the ground truth gestures ${Y}_{Pose}$, and (2) the \textit{similarity} between $\widehat{Y}_{Pose}$ and ${Y}_{Pose}$. To assess the effectiveness of the Image Schema input modality, we perform an ablation study where we compare two variations of our model: (1) the full model of \textbf{\textit{META4}}, and (2) a modified version of \textbf{\textit{META4}} with the image schema input modality ablated. Additionally, we conduct a sensitivity analysis to evaluate the model's reliance on the image schema input and its robustness in handling variations. To assess the reliability of the \textbf{\textit{BERTIS}} component within the overall framework of \textbf{\textit{META4}}, we evaluate it separately, by assessing the robustness and accuracy of this module in classifying text into Image Schemas.

\subsubsection{Metrics} 
For our ablation studies, we use the following metrics to assess the performance of \textbf{\textit{META4}}:
\begin{enumerate}
\item \textbf{Root Mean Squared Error (\textit{RMSE})} and \textbf{Mean Absolute Error (\textit{MAE})} are employed to measure the \textit{distance} between $\widehat{Y}_{Pose}$ and ${Y}_{Pose}$.
\item \textbf{Pearson Correlation Coefficient (\textit{PCC})} and \textbf{Cosine Similarity} are used to evaluate the \textit{similarity} between $\widehat{Y}_{Pose}$ and ${Y}_{Pose}$.
\end{enumerate}
To assess the classification accuracy of \textbf{\textit{BERTIS}} module in classifying text into image schemas, we use the following metrics:
\begin{enumerate}
\item \textbf{Precision}: measures the proportion of correctly predicted positive instances out of all instances predicted as positive. It focuses on the \textit{accuracy} of positive predictions.
\item \textbf{Recall}: measures the proportion of correctly predicted positive instances out of all actual positive instances. It focuses on the ability to capture positive instances.
\item \textbf{F1-score}: the harmonic mean of precision and recall, providing a balanced measure that combines both metrics
\end{enumerate}

\section{Results and Discussion}
\begin{table}[]
\centering
\begin{tabular}{@{}llll@{}}
\toprule
\textbf{Class}                                              & \textbf{Precision} & \textbf{Recall} & \textbf{F1-score} \\ \midrule \hline
"CENTER-PERIPHERY"                                                           & 0.98               & 0.94            & 0.96              \\ \midrule
"CONTACT"                                                           & 1                  & 1               & 1                 \\ \midrule
"CONTAINMENT"                                                           & 0.74               & 0.67            & 0.7               \\ \midrule
"COVERING"                                                           & 1                  & 1               & 1                 \\ \midrule
"FORCE"                                                           & 0.8                & 0.91            & 0.85              \\ \midrule
"LINK"                                                           & 1                  & 1               & 1                 \\ \midrule
"OBJECT"                                                           & 0.81               & 0.83            & 0.82              \\ \midrule
"PART-WHOLE"                                                           & 1                  & 1               & 1                 \\ \midrule
"SCALE"                                                           & 1                  & 1               & 1                 \\ \midrule
"SOURCE\_PATH\_GOAL"                                                           & 0.81               & 0.74            & 0.77              \\ \midrule
"SPLITTING"                                                          & 1                  & 1               & 1                 \\ \midrule
"SUBSTANCE"                                                          & 1                  & 1               & 1                 \\ \midrule
"SUPPORT"                                                          & 1                  & 1               & 1                 \\ \midrule
"VERTICALITY"                                                          & 0.88               & 0.93            & 0.90              \\ \midrule
\hline
\begin{tabular}[c]{@{}l@{}}Overall \\ Accuracy\end{tabular} & 0.93               &                 &                   \\ \bottomrule
\end{tabular}
\caption{Objective evaluation results of \textbf{\textit{BERTIS}}. Precision, Recall, and F1-score are computed for each of the 14 Image Schema classes. The overall accuracy is also computed. Higher scores indicate better performance for all three metrics.}
\label{BERTIS_eval}
\end{table}
\begin{figure*}[h]
  \centering
  \includegraphics[width=\linewidth]{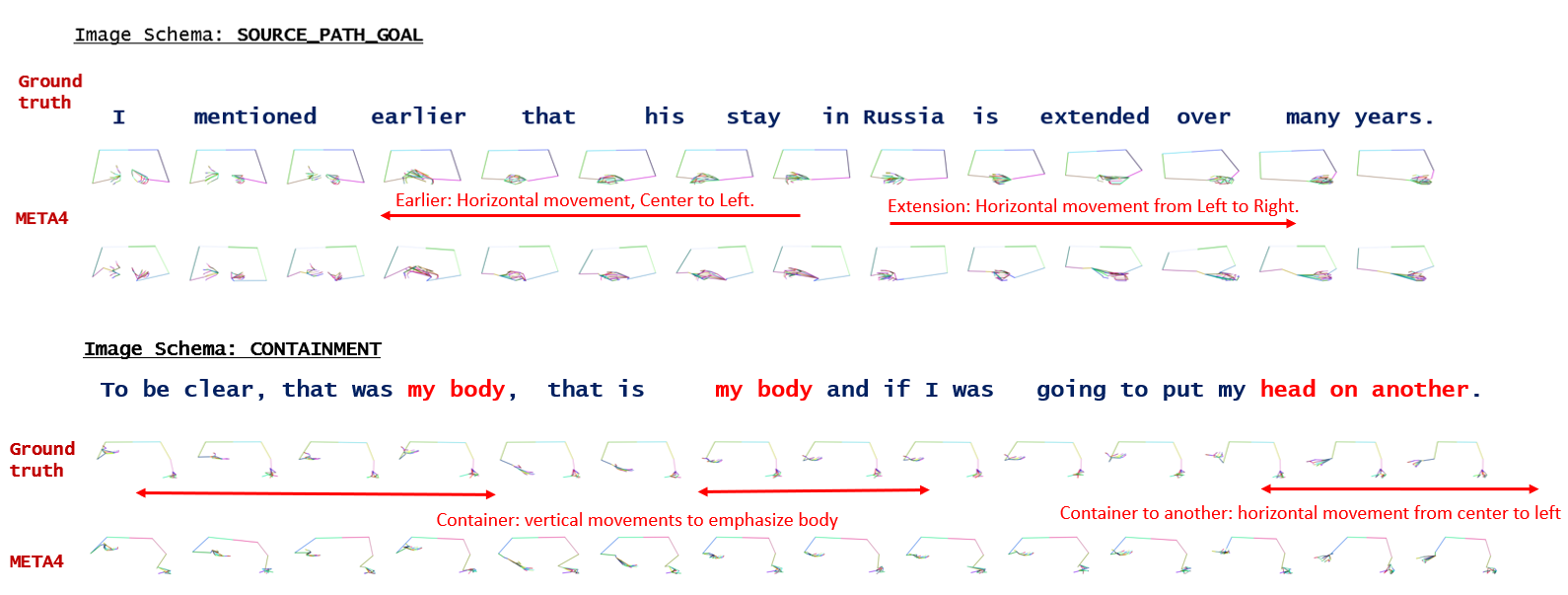}
  \caption{Visualizations of the Ground Truth Vs. META4 gestures for two utterances, one having a "SOURCE\_PATH\_GOAL" image schema, and the other a "CONTAINMENT" image schema. Fingers are added for sake of visualisation.}
  \label{post_hoc_visualization}
\end{figure*}
Table \ref{BERTIS_eval} reports the objective evaluation results of the evaluation conducted on \textbf{\textit{BERTIS}} for assessing its reliability in the framework of \textbf{\textit{META4}}, and its accuracy in classifying text into Image Schemas. For all Image Schema classes, \textit{F1-score} is between $0.77$ and $1$, indicating that \textbf{\textit{BERTIS}} performs well in terms of both accurately predicting the correct Image Schema classes (\textit{precision}) and capturing many correct Image Schema classes as possible (\textit{recall}). Indeed these results are also reflected in the \textit{precision} and \textit{recall} scores for all Image Schema classes. We observe a high \textit{precision} (between $0.74$ and $1$) for all classes, which indicates the number of correct predicted Image Schema classes. The \textit{recall} scores are above $0.74$ for most of the classes, with one Image Schema class having a \textit{recall} score equal to $0.67$. As reported in Table \ref{BERTIS_eval}, the overall \textit{accuracy} of \textbf{\textit{BERTIS}} with respect to all Image Schema classes is equal to $0.93$, indicating a high accuracy and good performance in classifying the input text into the correct Image Schema classes.
\begin{table}[H]
\tiny
\centering
\resizebox{\columnwidth}{!}{%
\begin{tabular}{llllll}
\hline
\multicolumn{2}{l}{} &
  \multicolumn{2}{c}{\textit{\textbf{Distance Metrics}}} &
  \multicolumn{2}{c}{\textit{\textbf{Similiarity Metrics}}} \\ \hline
\multicolumn{2}{c}{{\ul \textbf{Condition}}} &
  {\ul \textbf{RMSE}} &
  {\ul \textbf{MAE}} &
  {\ul \textbf{PCC}} &
  \textbf{\begin{tabular}[c]{@{}l@{}}Cosine \\ Similiarity\end{tabular}} \\ \hline \hline
\multirow{3}{*}{\textit{\textbf{\begin{tabular}[c]{@{}l@{}}Speaker \\ Dependent\end{tabular}}}} &
  \textbf{Full Model} &
  \textbf{0.01627} &
  \textbf{0.01197} &
  \textbf{0.98292} &
  \textbf{0.9985} \\ \cline{2-6} 
 & \textbf{IS Ablation} & 0.02004 & 0.01425 & 0.97645 & 0.997775 \\ \cline{2-6} 
 & \textbf{Mismatched}  & 0.01742 & 0.01222 & 0.96311 & 0.98415 \\ \hline
\multirow{3}{*}{\textit{\textbf{\begin{tabular}[c]{@{}l@{}}Speaker\\  Independent\end{tabular}}}} &
  \textbf{Full Model} &
  \textbf{0.02100} &
  \textbf{0.01543} &
  \textbf{0.97736} &
  \textbf{0.997497} \\ \cline{2-6} 
 & \textbf{IS Ablation} & 0.02520 & 0.01824 & 0.96779 & 0.99647  \\ \cline{2-6} 
 & \textbf{Mismatched}  & 0.02591 & 0.0161  & 0.97311 & 0.97313 \\ \hline
\end{tabular}%
}
\caption{Objective evaluation results of \textbf{\textit{META4}} for both conditions Speaker dependent and Speaker Independent. Results are reported for three conditions: the full model, the model with Image Schema (IS) modality ablation, and the mismatched condition.}
\label{objective_eval_results}
\end{table}

Table \ref{objective_eval_results} reports the ablation study as well as the sensitivity analysis results for both conditions \textit{Speaker Dependent (SD)} and \textit{Speaker Independent (SI)}. The full model performs best for both conditions in terms of errors produced w.r.t the Ground Truth (GT), and in terms of \textbf{\textit{META4}} predictions' correlation w.r.t to GT. When ablating the Image Schema modality, the \textbf{\textit{RMSE}} and \textbf{\textit{MAE}} errors increase for both \textit{SD} and \textit{SI} conditions. We notice a decrease in \textbf{PCC} and \textit{Cosine Similarity}, which indicate that the predictions became less correlated with the GT. These results confirm our hypothesis that Image Schema modality influences the predictions and therefore constitutes an important modality to consider when designing generative models for gesture synthesis. The sensitivity analysis results are also reported in Table \ref{objective_eval_results} under the condition "mismatched", which represents the error condition, and serves as a control condition to evaluate the robustness of \textbf{\textit{META4}} in handling Image Schema variations, allowing us to understand further the reliance of \textbf{\textit{META4}} on Image Schemas modality. For both conditions \textit{SD} and \textit{SI}, we notice an increase in terms \textit{RMSE} and \textit{MAE} errors, indicating that distorting the input Image Schema modality generates errors. The same conclusion can be drawn by looking at the \textit{PCC} and \textit{Cosine Similarity} between predictions and the ground truth, which decreased. 
Furthermore, looking at the results of \textit{SI} condition, we can validate \textbf{\textit{META4's}} ability to generalize its predictions to speakers that were not encountered during the training phase. 

\section{Post-Hoc Animation Visualization}
Figure \ref{post_hoc_visualization} illustrates the motion of a speaker saying two different utterances, each having a different Image Schema. The first key-frame animation corresponds to a text with a "SOURCE\_PATH\_GOAL" Image Schema. The speaker moves both of his arms from left to right. The same motion is reproduced by \textbf{\textit{META4}}. The second key-frame animation corresponds to a speaker saying an utterance with the image schema "CONTAINMENT". The speaker performs a vertical movement with his left hand to visually illustrate the concept of his body which he emphasizes in his utterance. He visually illustrate the metaphor of putting his head on another, he performs a horizontal movement.  \textbf{\textit{META4}} synthesizes the same behavior. This post-hoc visualization allows us to validate that \textbf{\textit{META4}} can reproduce the motion in the Ground Truth for different Image Schemas.

\section{Conclusion}
In this work, we propose an approach for synthesizing speech-driven metaphoric gestures while leveraging the potential of \textit{Image Schemas}. We conduct evaluation studies to demonstrate the effectiveness of our approach and show the importance of considering both speech and Image Schemas for generating \textit{metaphoric} gestures. In future work, we plan to conduct other evaluations to validate our approach subjectively on an Embodied Conversational Agent.

\newpage
{
{\LARGE\bfseries \underline{Appendix}}
}

\section*{Image Schema Classes}
The image schema classes that we consider in our work are 14, and are listed in Table \ref{IS_classes}.

\begin{table}[H]
\small
\centering
\small
\small
\begin{tabular}{|l|l|}
\hline

\multicolumn{1}{|c|}{\textbf{Image Schema Class}} & \multicolumn{1}{c|}{\textbf{Example}}                    \\ \hline
CENTER-PERIPHERY   & She brushed the thought away.            \\ \hline
CONTACT            & That blew me away.                       \\ \hline
CONTAINMENT        & Keep it in the back of your mind         \\ \hline
COVERING           & His judgement is clouded.                \\ \hline
FORCE              & They are attracted to each other.        \\ \hline
LINK               & breaking social ties                     \\ \hline
OBJECT             & Seize the opportunity.                   \\ \hline
PART-WHOLE         & They assembled a theory.                 \\ \hline
SCALE              & This class is bigger than that one.      \\ \hline
SOURCE\_PATH\_GOAL & The time for action has arrived.         \\ \hline
SPLITTING          & What separates the men from the boys?    \\ \hline
SUBSTANCE          & Emotions are tinged with suffuse         \\ \hline
VERTICALITY                                       & No known spoken language uses the lateral axis for time. \\ \hline
SUPPORT            & The poor in our country need a boost up. \\ \hline
\end{tabular}%
\caption{Image Schema classes with examples taken from \cite{wachowiak2022systematic}. }
\label{IS_classes}
\end{table}

\bibliographystyle{ACM-Reference-Format}
\bibliography{references} 

\end{document}